\pdfoutput=1
\def\year{2021}\relax
\documentclass[letterpaper]{article} 
\usepackage{aaai21}  
\usepackage{times}  
\usepackage{helvet} 
\usepackage{courier}  
\usepackage[hyphens]{url}  
\usepackage{graphicx} 
\usepackage{natbib}  
\usepackage{caption} 
\usepackage{xcolor}
\newcommand{\crossmark}{\textit{\sffamily x}}

\usepackage{booktabs}
\usepackage{amssymb}
\usepackage{multirow}
\usepackage{stmaryrd}
\title{A Generalization of Transformer Networks to Graphs}
\author {
    Vijay Prakash Dwivedi,\textsuperscript{\rm ¶} 
    Xavier Bresson\textsuperscript{\rm ¶} \\
}
\affiliations {
    \textsuperscript{\rm ¶} School of Computer Science and Engineering, Nanyang Technological University, Singapore\\
    \texttt{vijaypra001@e.ntu.edu.sg},  \texttt{xbresson@ntu.edu.sg}
}

\begin{document}

\maketitle

\begin{abstract}

We propose a generalization of transformer neural network architecture for arbitrary graphs.
The original transformer was designed for Natural Language Processing (NLP), which operates on fully connected graphs representing all connections between the words in a sequence. Such architecture does not leverage the graph connectivity inductive bias, and can perform poorly when the graph topology is important and has not been encoded into the node features. 
We introduce a graph transformer with four new properties compared to the standard model. 
First, the attention mechanism is a function of the neighborhood connectivity for each node in the graph. Second, the positional encoding is represented by the Laplacian eigenvectors, which naturally generalize the sinusoidal positional encodings often used in NLP. 
Third, the layer normalization is replaced by a batch normalization layer, which provides faster training and better generalization performance.
Finally, the architecture is extended to edge feature representation, which can be critical to tasks s.a. chemistry (bond type) or link prediction (entity relationship in knowledge graphs).
Numerical experiments on a graph benchmark demonstrate the performance of the proposed graph transformer architecture. 
This work closes the gap between the original transformer, which was designed for the limited case of line graphs, and graph neural networks, that can work with arbitrary graphs. 
As our architecture is simple and generic, we believe it can be used as a black box for future applications that wish to consider transformer and graphs.\footnote{\url{https://github.com/graphdeeplearning/graphtransformer}.}

\end{abstract}

\section{Introduction}

\begin{figure*}[!t]
\centering
\includegraphics[width=0.95\textwidth]{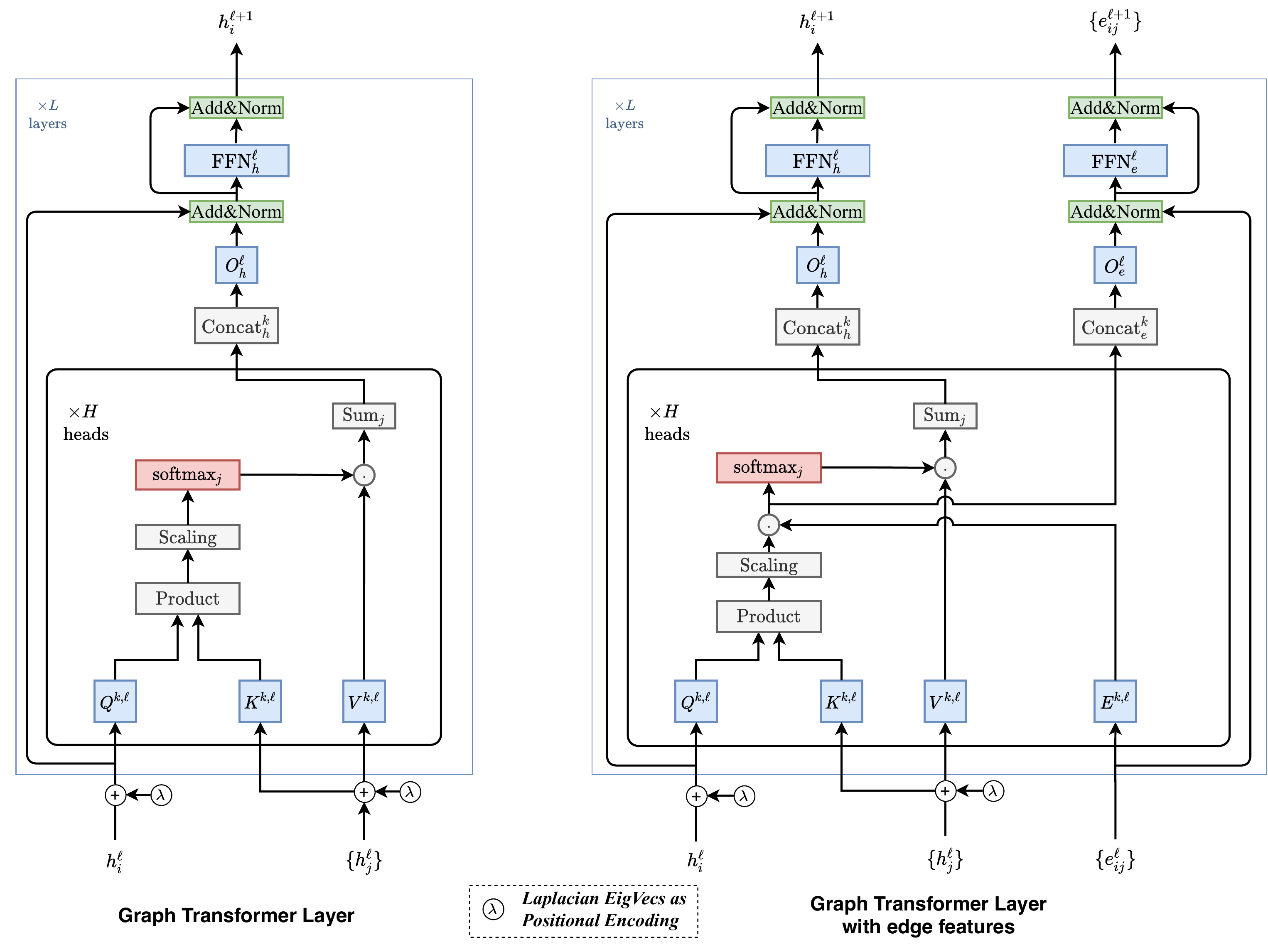} 
\caption{Block Diagram of Graph Transformer with Laplacian Eigvectors ($\lambda$) used as positional encoding (LapPE). LapPE is added to input node embeddings before passing the features to the first layer. \textit{\textbf{Left}}: Graph Transformer operating on node embeddings only to compute attention scores; \textit{\textbf{Right}}: Graph Transformer with edge features with designated feature pipeline to maintain layer wise edge representations. In this extension, the available edge attributes in a graph is used to explicitly modify the corresponding pairwise attention scores.}
\label{gt-architecture}
\end{figure*}

\noindent There has been a tremendous success in the field of natural language processing (NLP) since the development of Transformers \cite{vaswani2017attention} which are currently the best performing neural network architectures for handling long-term sequential datasets such as sentences in NLP. This is achieved by the use of attention mechanism \cite{bahdanau2014neural} where a word in a sentence attends to each other word and combines the received information to generate its abstract feature representations. From a perspective of message-passing paradigm \cite{gilmer2017neural} in graph neural networks (GNNs), this process of learning word feature representations by combining feature information from other words in a sentence can alternatively be viewed as a case of a GNN applied on a fully connected graph of words \cite{joshi2020transformers}. Transformers based models have led to state-of-the-art performance on several NLP applications \cite{devlin2018bert, radford2018improving, brown2020language}.
On the other hand, graph neural networks (GNNs) are shown to be the most effective neural network architectures on graph datasets and have achieved significant success on a wide range of applications, such as in knowledge graphs \cite{schlichtkrull2018modeling,chami2020low}, in social sciences \cite{monti2019fake}, in physics \cite{cranmer2019learning,sanchez2020learning}, etc. In particular, GNNs exploit the given arbitrary graph structure while learning the feature representations for nodes and edges and eventually the learned representations are used for downstream tasks. In this work, we explore inductive biases at the convergence of these two active research areas in deep learning towards presenting an improved version of Graph Transformer (see Figure \ref{gt-architecture}) which extends the key design components of the NLP transformers to arbitrary graphs.

\subsection{Related Work}
\label{sec:related_work}

As a preliminary, we highlight
the most recent research works 
which attempt
to develop graph transformers \cite{li2019graph, Nguyen2019U2GNN, zhang2020graph} with few focused on specialized cases such as on heterogeneous graphs, temporal networks, generative modeling, etc. \cite{yun2019graph, xu2019multi, hu2020heterogeneous, zhou2020data}. 

The model proposed in \citet{li2019graph} employs attention to all graph nodes instead of a node's local neighbors for the purpose of capturing global information. This limits the efficient exploitation of \textit{sparsity} which we show is a good inductive bias for learning on graph datasets. For the purpose of global information, we argue that there are other ways to incorporate the same instead of letting go sparsity and local contexts. For example, the use of graph-specific positional features \cite{zhang2020graph}, or node Laplacian position eigenvectors \cite{belkin2003laplacian, dwivedi2020benchmarking}, or relative learnable positional information \cite{you2019position}, virtual nodes \cite{li2015gated}, etc. 
\citet{zhang2020graph} propose Graph-BERT with an emphasis on pre-training and parallelized learning using a subgraph batching scheme that creates fixed-size linkless subgraphs to be passed to the model instead of the original graph. Graph-BERT employs a combination of several positional encoding schemes to capture absolute node structural and relative node positional information.
Since the original graph is not used \textit{directly} in Graph-BERT and the subgraphs do not have edges between the nodes (\textit{i.e.}, linkless), 
the proposed combination of positional encodings \textit{attempts at retaining} the original graph structure information in the nodes. 
We perform detailed analysis of Graph-BERT positional encoding schemes, along with experimental comparison with the model we present in this paper in Section \ref{sec:graph_bert_comparison}.

\citet{yun2019graph} developed Graph Transformer Networks (GTN) to learn on heterogeneous graphs with a target to transform a given heterogeneous graph into a meta-path based graph and then perform convolution. Notably, their focus behind the use of attention framework is for interpreting the generated meta-paths.
There is another transformer based approach developed for heterogeneous information networks, namely Heterogeneous Graph Transformer (HGT) by \citet{hu2020heterogeneous}. Apart from its ability of handling arbitrary number of node and edge types, HGT also captures the dynamics of information flow in the heterogeneous graphs in the form of relative temporal positional encoding which is based on the timestamp differences of the central node and the message-passing nodes. Furthermore, \citet{zhou2020data} proposed a transformer based generative model which generates temporal graphs by directly learning from dynamic information in networks.
The architecture presented in \citet{Nguyen2019U2GNN} somewhat proceeds along our goal to develop graph transformer for arbitrary homogeneous graphs with a coordinate embedding based positional encoding scheme. However, their experiments show that the coordinate embeddings are not universal in performance and only helps in a couple of unsupervised learning experiments among all evaluations. 

\subsection{Contributions}
Overall, we find that the most fruitful ideas from the transformers literature in NLP can be applied in a more efficient way and posit that sparsity and positional encodings are two \textit{key} aspects in the development of a Graph Transformer. 
As opposed to designing a best performing model for specific graph tasks, our work attempts for a generic, competitive transformer model which draws ideas together from the domains of NLP and GNNs.
For an overview, this paper brings the following contributions:
\begin{itemize}
    \item We put forward a generalization of transformer networks to homogeneous graphs of arbitrary structure, namely Graph Transformer, and an extended version of Graph Transformer with edge features that allows the usage of explicit domain information as edge features. 
    \item Our method includes an elegant way to fuse node positional features using Laplacian eigenvectors for graph datasets, inspired from the heavy usage of positional encodings in NLP transformer models and recent research on node positional features in GNNs.
    The comparison with literature shows Laplacian eigenvectors to be well-placed than any existing approaches to encode node positional information for arbitrary homogeneous graphs.
    \item Our experiments demonstrate that the proposed model surpasses baseline isotropic and anisotropic GNNs. The architecture simultaneously emerges as a better attention based GNN baseline as well as a simple and effective Transformer network baseline for graph datasets for future research at the intersection of attention and graphs.
\end{itemize}

\section{Proposed Architecture}
As stated earlier, we take into account two key aspects to develop Graph Transformers -- sparsity and positional encodings which should ideally be used in the best possible way for learning on graph datasets. We first discuss the motivations behind these using a transition from NLP to graphs, and then introduce the architecture proposed.

\subsection{On Graph Sparsity} In NLP transformers, a sentence is treated as a fully connected graph and this choice can be justified for two reasons -- a) First, it is \textit{difficult} to find meaningful sparse interactions or connections among the words in a sentence. For instance, the dependency of a word in a sentence on another word can vary with context, perspective of a user and specific application. There can be numerous plausible ground truth connections among words in a sentence and therefore, text datasets of sentences do not have explicit word interactions available. It thereby makes sense to have each word attending to each other word in a sentence, as followed by the Transformer architecture \cite{vaswani2017attention}.
-- b) Next, the so-called graph considered in an NLP transformer often has less than tens or hundreds of nodes (\textit{i.e.} sentences are often less than tens or hundreds of words). This makes for computationally feasibility and large transformer models can be trained on such fully connected graphs of words. 

In case of actual graph datasets, graphs have arbitrary connectivity structure available depending on the domain and target of application, and have node sizes in ranges of up to millions, or billions. The available structure presents us with a rich source of information to exploit as an inductive bias in a neural network, whereas the node sizes practically makes it impossible to have a fully connected graph for such datasets. On these accounts, it is ideal and practical to have a Graph Transformer where a node attends to local node neighbors, same as in GNNs \cite{NIPS2016_6081, kipf2017semi, Monti_2017, gilmer2017neural, velickovic2018graph, bresson2017residual, xu2018how}.

\subsection{On Positional Encodings}
\label{sec:on_positional_encodings}
In NLP, transformer based models are, in most cases, supplied with a positional encoding for each word. This is critical to ensure unique representation for each word, and eventually preserve distance information. For graphs, the design of unique node positions is challenging as there are symmetries which prevent canonical node positional information \cite{murphy2019relational}. In fact, most of the GNNs which are trained on graph datasets learn structural node information that are invariant to the node position \cite{srinivasan2019equivalence}. This is a critical reason why simple attention based models, such as GAT \cite{velickovic2018graph}, where the attention is a function of local neighborhood connectivity, instead full-graph connectivity, do not seem to achieve competitive performance on graph datasets. The issue of positional embeddings has been explored in recent GNN works \cite{murphy2019relational, you2019position, srinivasan2019equivalence, dwivedi2020benchmarking, li2020distance} with a goal to learn both structural and positional features. In particular, \citet{dwivedi2020benchmarking} make the use of available graph structure to pre-compute Laplacian eigenvectors \cite{belkin2003laplacian} and use them as node positional information. 
Since Laplacian PEs are generalization of the PE used in the original transformers \cite{vaswani2017attention} to graphs and these better help encode distance-aware information (\textit{i.e.,} nearby nodes have similar positional features and farther nodes have dissimilar positional features), we use Laplacian eigenvectors as PE in Graph Transformer. Although these eigenvectors have multiplicity occuring due to the arbitrary sign of eigenvectors, we randomly flip the sign of the eigenvectors during training, following \citet{dwivedi2020benchmarking}.
We pre-compute the Laplacian eigenvectors of all graphs in the dataset. Eigenvectors are defined via the factorization of the graph Laplacian matrix;
\begin{equation}
\Delta=\textrm{I}-D^{-1/2}AD^{-1/2}=U^T\Lambda U,
\label{eigenvec}
\end{equation}
where $A$ is the $n\times n$  adjacency matrix, $D$ is the degree matrix, and $\Lambda$, $U$ correspond to the eigenvalues and eigenvectors respectively. We use the $k$ smallest non-trivial eigenvectors of a node as its positional encoding and denote by ${\lambda}_i$ for node $i$. 
Finally, we refer to Section \ref{sec:graph_bert_comparison} for a comparison of Laplacian PE with existing Graph-BERT PEs.

\subsection{Graph Transformer Architecture}
We now introduce the Graph Transformer Layer and Graph Transformer Layer with edge features. The layer architecture is illustrated in Figure \ref{gt-architecture}. The first model is designed for graphs which do not have explicit edge attributes, whereas the second model maintains a designated edge feature pipeline to incorporate the available edge information and maintain their abstract representations at every layer.

\subsubsection{Input} First of all, we prepare the input node and edge embeddings to be passed to the Graph Transformer Layer. For a graph $\mathcal{G}$ with node features $\alpha_i \in \mathbb{R}^{d_n \times 1}$ for each node $i$ and edge features $\beta_{ij} \in \mathbb{R}^{d_e \times 1}$ for each edge between node $i$ and node $j$, the input node features $\alpha_i$ and edge features $\beta_{ij}$ are passed via a linear projection to embed these to $d$-dimensional hidden features $h_i^{0}$ and $e_{ij}^{0}$.
\begin{equation}
    \label{eqn:input_embd}
    \hat{h}_i^{0} = A^{0} \alpha_i + a^{0} \;\ ; \;\ e_{ij}^{0} = B^{0} \beta_{ij} + b^{0} ,
\end{equation}
where $A^{0} \in \mathbb{R}^{d \times d_n}$, $B^{0} \in \mathbb{R}^{d \times d_e}$ and $a^{0},b^{0}\in \mathbb{R}^{d}$ 
are the parameters of the linear projection layers.
We now embed the pre-computed node positional encodings of dim $k$ via a linear projection and add to the node features $\hat{h}_i^{0}$.
\begin{equation}
\label{eqn:pe_embd_add}
{\lambda}_i^{0} = C^{0} \lambda_i + c^{0} \;\  ; \;\ h_i^{0} = \hat{h}_i^{0} + {\lambda}_i^{0},  
\end{equation}
where $C^{0} \in \mathbb{R}^{d \times k}$ and $c^{0}\in \mathbb{R}^{d}$. Note that the Laplacian positional encodings are only added to the node features at the input layer and not during intermediate Graph Transformer layers.

\subsubsection{Graph Transformer Layer} The Graph Transformer is closely the same transformer architecture initially proposed in \cite{vaswani2017attention}, see Figure \ref{gt-architecture} (Left).
We now proceed to define the node update equations for a layer $\ell$. 

\begin{equation}
    \label{eqn:gt_layer}
    \hat{h}_{i}^{\ell+1} = O_h^{\ell} \ \bigparallel_{k=1}^{H} \Big(\sum_{j \in \mathcal{N}_i} w_{ij}^{k,\ell} V^{k,\ell}h_j^{\ell} \Big),
\end{equation}
\begin{equation}
    \label{eqn:softmax}
    \textnormal{where,} \ w_{ij}^{k,\ell} = \textnormal{softmax}_j \Big(\frac{Q^{k, \ell} h_i^{\ell} \ \cdot \ K^{k, \ell}h_j^{\ell}}{\sqrt{d_k}}  \Big),
\end{equation}
and $Q^{k,\ell}, K^{k,\ell}, V^{k,\ell} \in \mathbb{R}^{d_k \times d}$, $O_h^{\ell} \in \mathbb{R}^{d \times d}$, $k=1$ to $H$ denotes the number of attention heads, and $\|$ denotes concatenation. 
For numerical stability, the outputs after taking exponents of the terms inside softmax is clamped to a value between $-5$ to $+5$.
The attention outputs $\hat{h}_{i}^{\ell+1}$ are then passed to a Feed Forward Network (FFN) preceded and succeeded by residual connections and normalization layers, as:

\begin{eqnarray}
    \hat{\hat{h}}_{i}^{\ell+1} &=& \textnormal{Norm} \Big( h_{i}^{\ell} + \hat{h}_{i}^{\ell+1} \Big), \label{eqn:rc_norm1}\\
    \hat{\hat{\hat{h}}}_{i}^{\ell+1} &=& W_2^{\ell} \textnormal{ReLU}(W_1^{\ell} \hat{\hat{h}}_{i}^{\ell+1}), \label{eqn:ffn}\\
    h_{i}^{\ell+1} &=& \textnormal{Norm} \Big( \hat{\hat{h}}_{i}^{\ell+1} + \hat{\hat{\hat{h}}}_{i}^{\ell+1} \Big), \label{eqn:rc_norm2}
\end{eqnarray}
where $W_1^{\ell}, \in \mathbb{R}^{2d \times d}$, $W_2^{\ell}, \in \mathbb{R}^{d \times 2d}$, $\hat{\hat{h}}_{i}^{\ell+1}, \hat{\hat{\hat{h}}}_{i}^{\ell+1}$ denote intermediate representations, and  Norm can either be LayerNorm\cite{ba2016layer} or BatchNorm \cite{ioffe2015batch}. The bias terms are omitted for clarity of presentation.

\subsubsection{Graph Transformer Layer with edge features} The Graph Transformer with edge features is designed for better utilization of rich feature information available in several graph datasets in the form of edge attributes. See Figure \ref{gt-architecture} (Right) for a reference to the building block of a layer. Since our objective remains to better use the edge features which are pairwise scores corresponding to a node pair, we tie these available edge features to implicit edge scores computed by pairwise attention. In other words, say an intermediate attention score before softmax, $\hat{w}_{ij}$, is computed when a node $i$ attends to node $j$ after the 
multiplication of \textit{query} and \textit{key} feature projections, see the expression inside the brackets in Equation \ref{eqn:softmax}. Let us treat this score $\hat{w}_{ij}$ as implicit information about the edge $<i,j>$. We now try to inject the available edge information for the edge $<i,j>$ and improve the already computed implicit attention score $\hat{w}_{ij}$. It is done by simply multiplying the two values $\hat{w}_{ij}$ and $e_{ij}$, see Equation \ref{eqn:softmax_edge_2}. This kind of information injection is not seen to be explored much, or applied in NLP Transformers as there is usually no available feature information between two words. However, in graph datasets such as molecular graphs, or social media graphs, there is often some feature information available on the edge interactions and it becomes natural to design an architecture to use this information while learning. For the edges,
we also maintain a designated node-symmetric edge feature representation pipeline for propagating edge attributes from one layer to another, see Figure \ref{gt-architecture}.
We now proceed to define the layer update equations for a layer $\ell$. 

\begin{eqnarray}
    \hat{h}_{i}^{\ell+1} &=& O_h^{\ell} \ \bigparallel_{k=1}^{H} \Big(\sum_{j \in \mathcal{N}_i} w_{ij}^{k,\ell} V^{k,\ell}h_j^{\ell} \Big), \label{eqn:gt_layer_edge_h}\\
    \hat{e}_{ij}^{\ell+1} &=& O_e^{\ell} \ \bigparallel_{k=1}^{H} \Big(\hat{w}_{ij}^{k,\ell} \Big), \ \ \textnormal{where,} \label{eqn:gt_layer_edge_e}
\end{eqnarray}
\begin{eqnarray}
    w_{ij}^{k,\ell} &=& \textnormal{softmax}_j ( \hat{w}_{ij}^{k,\ell} ), \label{eqn:softmax_edge_1}\\
    \hat{w}_{ij}^{k,\ell} &=& \Big( \frac{Q^{k, \ell} h_i^{\ell} \ \cdot \ K^{k, \ell}h_j^{\ell}}{\sqrt{d_k}} \Big)  \ \cdot \ E^{k,\ell} e^{\ell}_{ij} , \label{eqn:softmax_edge_2}
\end{eqnarray}
and $Q^{k,\ell}, K^{k,\ell}, V^{k,\ell}, E^{k,\ell} \in \mathbb{R}^{d_k \times d}$, $O_h^{\ell}, O_e^{\ell} \in \mathbb{R}^{d \times d}$, $k=1$ to $H$ denotes the number of attention head, and $\|$ denotes concatenation. 
For numerical stability, the outputs after taking exponents of the terms inside softmax is clamped to a value between $-5$ to $+5$.
The outputs $\hat{h}_{i}^{\ell+1}$ and $\hat{e}_{ij}^{\ell+1}$ are then passed to separate Feed Forward Networks preceded and succeeded by residual connections and normalization layers, as:

\begin{eqnarray}
    \hat{\hat{h}}_{i}^{\ell+1} &=& \textnormal{Norm} \Big( h_{i}^{\ell} + \hat{h}_{i}^{\ell+1} \Big), \label{eqn:rc_norm1_h}\\
    \hat{\hat{\hat{h}}}_{i}^{\ell+1} &=& W_{h,2}^{\ell} \textnormal{ReLU}(W_{h,1}^{\ell} \hat{\hat{h}}_{i}^{\ell+1}), \label{eqn:ffn_h}\\
    h_{i}^{\ell+1} &=& \textnormal{Norm} \Big( \hat{\hat{h}}_{i}^{\ell+1} + \hat{\hat{\hat{h}}}_{i}^{\ell+1} \Big), \label{eqn:rc_norm2_h}
\end{eqnarray}
where $W_{h,1}^{\ell}, \in \mathbb{R}^{2d \times d}$, $W_{h,2}^{\ell}, \in \mathbb{R}^{d \times 2d}$, $\hat{\hat{h}}_{i}^{\ell+1}, \hat{\hat{\hat{h}}}_{i}^{\ell+1}$ denote intermediate representations,
\begin{eqnarray}
    \hat{\hat{e}}_{ij}^{\ell+1} &=& \textnormal{Norm} \Big( e_{ij}^{\ell} + \hat{e}_{ij}^{\ell+1} \Big), \label{eqn:rc_norm1_e}\\
    \hat{\hat{\hat{e}}}_{ij}^{\ell+1} &=& W_{e,2}^{\ell} \textnormal{ReLU}(W_{e,1}^{\ell} \hat{\hat{e}}_{ij}^{\ell+1}), \label{eqn:ffn_e}\\
    e_{ij}^{\ell+1} &=& \textnormal{Norm} \Big( \hat{\hat{e}}_{ij}^{\ell+1} + \hat{\hat{\hat{e}}}_{ij}^{\ell+1} \Big), \label{eqn:rc_norm2_e}
\end{eqnarray}
where $W_{e,1}^{\ell}, \in \mathbb{R}^{2d \times d}$, $W_{e,2}^{\ell}, \in \mathbb{R}^{d \times 2d}$, $\hat{\hat{e}}_{ij}^{\ell+1}, \hat{\hat{\hat{e}}}_{ij}^{\ell+1}$ denote intermediate representations. 

\subsubsection{Task based MLP Layers} 
The node representations obtained at the final layer of Graph Transformer are passed to a task based MLP network for computing task-dependent outputs, which are then fed to a loss function to train the parameters of the model. The formal definitions of the task based layers that we use can be found in Appendix \ref{sec:task_based_layers}.

\section{Numerical Experiments}

\begin{table*}[t!]
    \centering
    \scalebox{0.70}{
    \begin{tabular}{rccc|cccc|cccc}
        \toprule
        & & & & \multicolumn{4}{c}{\textbf{Sparse Graph}} & \multicolumn{4}{|c}{\textbf{Full Graph}}\\
        \textbf{Dataset}  & \textbf{LapPE} & \textbf{$L$} & \textbf{\#Param} & \textbf{Test Perf.$\pm$s.d.} & \textbf{Train Perf.$\pm$s.d.} & \textbf{\#Epoch} & \textbf{Epoch/Total} & \textbf{Test Perf.$\pm$s.d.} & \textbf{Train Perf.$\pm$s.d.} & \textbf{\#Epoch} & \textbf{Epoch/Total} \\
        \midrule
        \midrule
        \multicolumn{12}{c}{Batch Norm: \texttt{False}; Layer Norm: \texttt{True}}\\
        \midrule
        \multirow{2}{*}{\textbf{ZINC}} & \crossmark & 10 & 588353 & 0.278$\pm$0.018 & 0.027$\pm$0.004 & 274.75 & 26.87s/2.06hr & 0.741$\pm$0.008 & 0.431$\pm$0.013 & 196.75 & 37.64s/2.09hr\\
        & \checkmark & 10 & 588929 & 0.284$\pm$0.012 & 0.031$\pm$0.006 & 263.00 & 26.64s/1.98hr & 0.735$\pm$0.006 & 0.442$\pm$0.031 & 196.75 & 31.50s/1.77hr\\
        \midrule
        \multirow{2}{*}{\textbf{CLUSTER}} & \crossmark & 10 & 523146 & 70.879$\pm$0.295 & 86.174$\pm$0.365 & 128.50 & 202.68s/7.32hr & 19.596$\pm$2.071 & 19.570$\pm$2.053 & 103.00 & 512.34s/15.15hr\\
        & \checkmark & 10 & 524026 & 70.649$\pm$0.250 & 86.395$\pm$0.528 & 130.75 & 200.55s/7.43hr & 27.091$\pm$3.920 & 26.916$\pm$3.764 & 139.50 & 565.13s/22.37hr\\
        \midrule
        \multirow{2}{*}{\textbf{PATTERN}} & \crossmark & 10 & 522742 & 73.140$\pm$13.633 & 73.070$\pm$13.589 & 184.25 & 276.66s/13.75hr & 50.854$\pm$0.111 & 50.906$\pm$0.005 & 108.00 & 540.85s/16.77hr\\
        & \checkmark & 10 & 522982 & 71.005$\pm$11.831 & 71.125$\pm$11.977 & 192.50 & 294.91s/14.79hr & 56.482$\pm$3.549 & 56.565$\pm$3.546 & 124.50 & 637.55s/22.69hr\\
        \midrule
        \midrule
        \multicolumn{12}{c}{Batch Norm: \texttt{True}; Layer Norm: \texttt{False}}\\
        \midrule
        \multirow{2}{*}{\textbf{ZINC}} & \crossmark & 10 & 588353 & 0.264$\pm$0.008 & 0.048$\pm$0.006 & 321.50 & 28.01s/2.52hr & 0.724$\pm$0.013 & 0.518$\pm$0.013 & 192.25 & 50.27s/2.72hr\\
        & \checkmark & 10 & 588929 & \textbf{0.226$\pm$0.014} & 0.059$\pm$0.011 & 287.50 & 27.78s/2.25hr & 0.598$\pm$0.049 & 0.339$\pm$0.123 & 273.50 & 45.26s/3.50hr\\
        \midrule
        \multirow{2}{*}{\textbf{CLUSTER}} & \crossmark & 10 & 523146 & 72.139$\pm$0.405 & 85.857$\pm$0.555 & 121.75 & 200.85s/6.88hr & 21.092$\pm$0.134 & 21.071$\pm$0.037 & 100.25 & 595.24s/17.10hr\\
        & \checkmark & 10 & 524026 & \textbf{73.169$\pm$0.622} & 86.585$\pm$0.905 & 126.50 & 201.06s/7.20hr & 27.121$\pm$8.471 & 27.192$\pm$8.485 & 133.75 & 552.06s/20.72hr\\
        \midrule
        \multirow{2}{*}{\textbf{PATTERN}} & \crossmark & 10 & 522742 & 83.949$\pm$0.303 & 83.864$\pm$0.489 & 236.50 & 299.54s/19.71hr & 50.889$\pm$0.069 & 50.873$\pm$0.039 & 104.50 & 621.33s/17.53hr\\
        & \checkmark & 10 & 522982 & \textbf{84.808$\pm$0.068} & 86.559$\pm$0.116 & 145.25 & 309.95s/12.67hr & 54.941$\pm$3.739 & 54.915$\pm$3.769 & 117.75 & 683.53s/22.77hr\\
        \bottomrule
        \end{tabular}
    }
    \caption{
    Results of GraphTransformer (GT) on all datasets. Performance Measure for ZINC is MAE, for PATTERN and CLUSTER is Acc. Results (higher is better for all except ZINC) are averaged over 4 runs with 4 different seeds. \textbf{Bold}: the best performing model for each dataset. We perform each experiment with given graphs \textbf{(Sparse Graph)} and \textbf{(Full Graph)} in which we create full connections among all nodes; For ZINC full graphs, edge features are discarded given our motive of the full graph experiments without any sparse structure information.
    }
    \label{tab:results}
\end{table*}

\begin{table}[t!]
    \centering
    \scalebox{0.80}{
    \begin{tabular}{rccc}
        \toprule
        \textbf{Model}  & \textbf{ZINC} & \textbf{CLUSTER} & \textbf{PATTERN} \\
        \midrule
        \multicolumn{4}{c}{\texttt{GNN BASELINE SCORES} from \cite{dwivedi2020benchmarking}}\\
        \midrule
        \textbf{GCN} & 0.367$\pm$0.011 & 68.498$\pm$0.976 & 71.892$\pm$0.334\\
        \textbf{GAT} & 0.384$\pm$0.007 & 70.587$\pm$0.447 & 78.271$\pm$0.186\\
        \textbf{GatedGCN} & 0.214$\pm$0.013 & 76.082$\pm$0.196 & 86.508$\pm$0.085 \\
        \midrule
        \multicolumn{4}{c}{\texttt{OUR RESULTS}}\\
        \midrule
        \textbf{GT (Ours)} & 0.226$\pm$0.014 & 73.169$\pm$0.622 & 84.808$\pm$0.068\\
        \bottomrule
        \end{tabular}
    }
    \caption{
    Comparison of our best performing scores (from Table \ref{tab:results}) on each dataset against the GNN baselines (GCN \cite{kipf2017semi}, GAT \cite{velickovic2018graph}, GatedGCN\cite{bresson2017residual}) 
    of 500k model parameters. 
    \textbf{Note:} Only GatedGCN and GT models use the available edge attributes in ZINC.
    }
    \label{tab:results_comparison}
\end{table}

We evaluate the performance of proposed Graph Transformer on three benchmark graph datasets-- ZINC \cite{irwin2012zinc}, PATTERN and CLUSTER \cite{abbe2017community} from a recent GNN benchmark \cite{dwivedi2020benchmarking}. 

\subsubsection{ZINC, Graph Regression}
ZINC \cite{irwin2012zinc} is a molecular dataset with the task of graph property regression for constrained solubility. Each ZINC molecule is represented as a graph of atoms as nodes and bonds as edges. Since this dataset have rich feature information in terms of bonds as edge attributes, we use the `Graph Transformer with edge features' for this task. We use the 12K subset of the data as in \citet{dwivedi2020benchmarking}.
\subsubsection{PATTERN, Node Classification}
PATTERN is a node classification dataset generated using the Stochastic Block Models (SBM) \cite{abbe2017community}. The task is classify the nodes into 2 communities. PATTERN graphs do not have explicit edge features and hence we use the simple `Graph Transformer' for this task. The size of this dataset is 14K graphs.
\subsubsection{CLUSTER, Node Classification}
CLUSTER is also a synthetically generated dataset using SBM model. The task is to assign a cluster label to each node. There are total 6 cluster labels. Similar to PATTERN, CLUSTER graphs do not have explicit edge features and hence we use the simple `Graph Transformer' for this task. The size of this dataset is 12K graphs. We refer the readers to \cite{dwivedi2020benchmarking} for additional information, inlcuding preparation, of these datasets.

\subsubsection{Model Configurations}
For experiments, we follow the benchmarking protocol introduced in \citet{dwivedi2020benchmarking} based on PyTorch \cite{paszke2019pytorch} and DGL \cite {wang2019dgl}. We use 10 layers of Graph Transformer layers with each layer having 8 attention heads and arbitrary hidden dimensions such that the total number of trainable parameters is in the range of 500k. We use learning rate decay strategy to train the models where the training stops at a point when the learning rate reaches to a value of $1 \times 10^{-6}$. 
We run each experiment with 4 different seeds and report the mean and average performance measure of the 4 runs. 
The results are reported in Table \ref{tab:results} and comparison in Table \ref{tab:results_comparison}.

\section{Analysis and Discussion}
We now present the analysis of  our experiments on the proposed Graph Transformer Architecture,
see Tables \ref{tab:results} and \ref{tab:results_comparison}.

\begin{itemize}
    \item The generalization of transformer network on graphs is best when Laplacian PE are used for node positions and Batch Normalization is selected instead of Layer Normalization. For all three benchmark datasets, the experiments score the highest performance in this setting, see Table \ref{tab:results}.
    \item The proposed architecture performs significantly better than baseline isotropic and anisotropic GNNs (GCN and GAT respectively), and helps close the gap between the original transformer and transformer for graphs. Notably, our architecture emerges as a fresh and improved attention based GNN baseline surpassing GAT (see Table \ref{tab:results_comparison}), which employs multi-headed attention inspired by the original transformer  \cite{vaswani2017attention} and have been often used in the literature as a baseline for attention-based GNN models.
    \item As expected, sparse graph connectivity is a critical inductive bias for datasets with arbitrary graph structure, as demonstrated by comparing sparse vs. full graph experiments. 
    \item Our proposed extension of Graph Transformer with edge features reaches close to the best performing GNN, \textit{i.e.,} GatedGCN, on ZINC. This architecture specifically brings exciting promise to datasets where domain information along pairwise interactions can be leveraged for maximum learning performance.
\end{itemize}

\subsection{Comparison to PEs used in Graph-BERT}
In addition to the 
reasons underscored in Sections \ref{sec:related_work} and \ref{sec:on_positional_encodings}, we demonstrate the usefulness of Laplacian eigenvectors as a suitable candidate PE for Graph Transformer in this section, by its comparison with different PE schemes applied in Graph-BERT \cite{zhang2020graph}.\footnote{Note that we do not perform empirical comparison with other PEs in Graph Transformer literature except Graph-BERT, because of two reasons: i) Some existing Graph Transformer methods do not use PEs, ii) If PEs are used, they are usually specialised; for instance, Relative Temporal Encoding (RTE) for encoding dynamic information in heterogeneous graphs in \cite{hu2020heterogeneous}.} In Graph-BERT, which operates on fixed size sampled subgraphs, a node attends to every other node in a subgraph. 
For a given graph $\mathcal{G}=(\mathcal{V},\mathcal{E})$ with $\mathcal{V}$ nodes and $\mathcal{E}$ edges, a subgraph $g_i$ of size $k+1$ is created for every node $i$ in the graph, which means the original single graph $\mathcal{G}$ is converted to $\mathcal{V}$ subgraphs. For a subgraph $g_i$ corresponding to node $u_i$, the $k$ other nodes are the ones which have the top $k$ intimacy scores with node $u_i$ based on a pre-computed intimacy matrix that maps every edge in the graph $\mathcal{G}$ to an intimacy score.
While the sampling is great for parallelization and efficiency, the original graph structure is not directly used in the layers. Graph-BERT uses a combination of node PE schemes to inform the model on node structural, positional, and distance information from original graph-- i) Intimacy based relative PE, ii) Hop based relative distance encoding, and iii) Weisfeiler Lehman based absolute PE (WL-PE). The intimacy based PE and the hop based PE are variant to the sampled subgraphs, \textit{i.e.}, these PEs for a node in a subgraph $g_i$ depends on the node $u_i$ w.r.t which it is sampled, and cannot be directly used in other cases unless we use similar sampling strategy. The WL-PE which are absolute structural roles of nodes in the original graph computed using WL algorithm \cite{zhang2020graph, niepert2016learning}, are not variant to the subgraphs and can be easily used as a generic PE mechanism. On that account, we swap Laplacian PE in our experiments for an ablation analysis and use WL-PE from Graph-BERT, see Table \ref{tab:results_ablation}.
As Laplacian PE capture better structural and positional information about the nodes, which essentially is the objective behind using the three Graph-BERT PEs, they outperform the WL-PE. Besides, WL-PEs tend to overfit SBM datasets and lead to poor generalization.

\label{sec:graph_bert_comparison}
\begin{table}[t!]
    \centering
    \scalebox{0.58}{
    \begin{tabular}{r|r|c|cccc}
        \toprule
        & & & \multicolumn{4}{c}{\textbf{Sparse Graph}} \\
        \textbf{Dataset} & \textbf{PE} & \textbf{\#Param} & \textbf{Test Perf.$\pm$s.d.} & \textbf{Train Perf.$\pm$s.d.} & \textbf{\#Epoch} & \textbf{Epoch/Total} \\
        \midrule
        \midrule
        \multicolumn{7}{c}{Batch Norm: \texttt{True}; Layer Norm: \texttt{False}; $L$ = 10}\\
        \midrule
        \multirow{3}{*}{\textbf{ZINC}} & \crossmark & 588353 & 0.264$\pm$0.008 & 0.048$\pm$0.006 & 321.50 & 28.01s/2.52hr \\
        & L & 588929 & \textbf{0.226$\pm$0.014} & 0.059$\pm$0.011 & 287.50 & 27.78s/2.25hr \\
        & W & 590721 & 0.267$\pm$0.012 & 0.059$\pm$0.010 & 263.25 & 27.04s/2.00hr\\
        \midrule
        \multirow{3}{*}{\textbf{CLUSTER}}& \crossmark & 523146 & 72.139$\pm$0.405 & 85.857$\pm$0.555 & 121.75 & 200.85s/6.88hr \\
        & L & 524026 & \textbf{73.169$\pm$0.622} & 86.585$\pm$0.905 & 126.50 & 201.06s/7.20hr \\
        & W & 531146 & 70.790$\pm$0.537 & 86.829$\pm$0.745 & 119.00 & 196.41s/6.69hr\\
        \midrule
        \multirow{3}{*}{\textbf{PATTERN}} & \crossmark & 522742 & 83.949$\pm$0.303 & 83.864$\pm$0.489 & 236.50 & 299.54s/19.71hr \\
        & L & 522982 & \textbf{84.808$\pm$0.068} & 86.559$\pm$0.116 & 145.25 & 309.95s/12.67hr \\
        & W & 530742 & 75.489$\pm$0.216 & 97.028$\pm$0.104 & 109.25 & 310.11s/9.73hr\\
        \bottomrule
        \end{tabular}
    }
    \caption{
    Analysis of GraphTransformer (GT) using different PE schemes. Notations \textbf{x}: No PE; \textbf{L}: LapPE (ours); \textbf{W}: WL-PE \cite{zhang2020graph}. 
    \textbf{Bold}: the best performing model for each dataset. 
    }
    \label{tab:results_ablation}
\end{table}

\section{Conclusion}
This work presented a simple yet effective approach to generalize transformer networks on arbitrary graphs and introduced the corresponding architecture. Our experiments consistently showed that the presence of -- i) Laplacian eigenvectors as node positional encodings and -- ii) batch normalization, in place of layer normalization, around the transformer feed forward layers enhanced the transformer universally on all experiments.
Given the simple and generic nature of our architecture and competitive performance against standard GNNs, we believe the proposed model can be used as baseline for further improvement across graph applications employing node attention. In future works, we are interested in building upon the graph transformer along aspects such as efficient training on single large graphs, applicability on heterogeneous domains, etc., and perform efficient graph representation learning keeping in account the recent innovations in graph inductive biases.

\section*{Acknowledgments}
XB is supported by NRF Fellowship NRFF2017-10.

\bibliography{bibfile}

\newpage
\appendix
\section{Appendix}

\subsection{Task based MLP layer equations}

\paragraph{Graph prediction layer}
\label{sec:task_based_layers}
For graph prediction task, the final layer node features of a graph is averaged to get a $d$-dimensional graph-level feature vector $y_\mathcal{G}$.

\begin{equation}
    \label{eqn:graph-repres}
    y_\mathcal{G} =  \frac{1}{\mathcal{V}} \sum_{i=0}^{\mathcal{V}}{h_{i}^{L}},
\end{equation}
The graph feature vector is then passed to a MLP to obtain the un-normalized prediction score for each class, $y_\textnormal{pred}\in \mathbb{R}^C$ for each class:
\begin{equation}
    \label{eqn:graph_prediction}
    y_\textnormal{pred} = P \ \textnormal{ReLU} \left( Q \ y_\mathcal{G} \right),
\end{equation}
where $P \in \mathbb{R}^{d \times C}, Q \in \mathbb{R}^{d \times d}, C$ is the number of task labels (classes) to be predicted. 
Since we perform single-target graph regression in ZINC, $C=1$, and the L1-loss between the predicted and groundtruth values is minimized during training.

\paragraph{Node prediction layer}
For node prediction task, each node's feature vector is passed to a MLP for computing the un-normalized prediction scores $y_{i,\textnormal{pred}}\in \mathbb{R}^C$ for each class:
\begin{equation}
    \label{eqn:node_prediction}
    y_{i,\textnormal{pred}} = P \ \textnormal{ReLU} \left( Q \ h_{i}^{L} \right),
\end{equation}
where $P \in \mathbb{R}^{d \times C}, Q \in \mathbb{R}^{d \times d}$. During training, the cross-entropy loss weighted inversely by the class size is used.\\

As a note, these task based layers can be modified as per the requirements of the dataset, and or the prediction to be done. For example, the Graph Transformer edge outputs (Figure \ref{gt-architecture} (Right)) can be used for edge prediction tasks and the task based MLP layers can be defined in similar fashion as we do for node prediction. Besides, different styles of using final and/or intermediate Graph Transformer layers can be used as inputs to the task based MLP layers, such as JK Readout (Jumping Knowledge) \cite{xu2018representation}, etc. used often in GNNs.

\subsection{Hardware Information}
All experiments are run on Intel Xeon CPU E5-2690 v4 server with 4 Nvidia 1080Ti GPUs. At a given time, 4 experiments were run on the server with each single GPU running 1 experiment.
The maximum training time for an experiment is limited to 24 hours.

\end{document}